\documentclass[review]{elsarticle}

\usepackage{lineno,hyperref}
\usepackage{url}            
\usepackage{booktabs}       
\usepackage{amsfonts}       
\usepackage{nicefrac}       
\usepackage{microtype}      
\usepackage{amsmath}
\usepackage{multirow}
\usepackage{amssymb}
\usepackage{algorithmic,algorithm}
\usepackage{graphicx}
\usepackage{wrapfig}
\usepackage{soul}
\usepackage[dvipsnames]{xcolor}
\modulolinenumbers[5]

\newtheorem{theorem}{Theorem}[section]

\journal{peer review}

\begin{document}

\begin{frontmatter}

\title{Bayesian Active Learning with Abstention Feedbacks}

\author[firstaddress]{Cuong V. Nguyen\corref{beforeorg}}
\author[secondaddress]{Lam Si Tung Ho}
\author[thirdaddress]{Huan Xu\corref{beforeorg}}
\author[fourthaddress]{Vu Dinh}
\author[seventhaddress,eighthaddress,ninthaddress]{Binh T. Nguyen\corref{correspondingauthor}}
\cortext[beforeorg]{Work done before joining current organizations.}
\cortext[correspondingauthor]{Corresponding author}
\ead{ngtbinh@hcmus.edu.vn}
\address[firstaddress]{Amazon Web Services, USA}
\address[secondaddress]{Dalhousie University, Dalhousie, Canada}
\address[thirdaddress]{Alibaba Inc., USA}
\address[fourthaddress]{University of Delaware, USA}
\address[seventhaddress]{AISIA Research Lab, Ho Chi Minh City, Vietnam}
\address[eighthaddress]{Department of Computer Science, University of Science, Ho Chi Minh City, Vietnam}
\address[ninthaddress]{Vietnam National University, Ho Chi Minh City, Vietnam}

\begin{abstract}
We study pool-based active learning with abstention feedbacks where a labeler can abstain from labeling a queried example with some unknown abstention rate.
This is an important problem with many useful applications.
We take a Bayesian approach to the problem and develop two new greedy algorithms that learn both the classification problem and the unknown abstention rate at the same time.
These are achieved by simply incorporating the estimated average abstention rate into the greedy criteria.
We prove that both algorithms have near-optimality guarantees: they respectively achieve a ${(1-\frac{1}{e})}$ constant factor approximation of the optimal expected or worst-case value of a useful utility function.
Our experiments show the algorithms perform well in various practical scenarios.
\end{abstract}


\end{frontmatter}


 \section{Introduction}
\label{sec:intro}

We consider the problem of active learning with abstention feedbacks where a labeler can abstain from labeling queried examples with some unknown abstention rate.
This problem is one of the several attempts to deal with imperfect labelers in active learning who may give incorrect or noisy labels to queried examples \citep{zhang2015active,yan2015active} or in our case, give abstention feedbacks to queries \citep{fang2012active,ramirez2014anytime,yan2015active}.

Learning with abstention feedbacks is important in many real-life scenarios.
Below we discuss some examples where this problem is useful.
In these examples, although the reasons for the abstention vary, from the learner's view they are the same:
the learner will receive no labels for some queries and the true labels for the others.

{\it Crowdsourcing:}
In crowdsourcing, we have many labelers, each of whom only has expertise in some certain area and therefore can only provide labels for a subset of the input domain.
These labelers were also called labelers with a knowledge blind spot \citep{fang2012active}.
In this case, active learning is a good approach to quickly narrow down the expertise domain of a labeler and focus on querying examples in this region to learn a good model.
By adapting active learning algorithms to each labeler, we can also gather representative subsets of labeled data from the labelers and combine them into a final training set.

{\it Learning with Corrupted Labels:}
In this problem, the abstention feedbacks do not come from the labeler but occur due to corruptions in the labels received by the learner.
The corruptions could be caused by bad communication channels that distort the labels or could even be caused by attackers attempting to corrupt the labels \citep{zhao2017efficient}.
The setting in our paper can deal with the case when the corrupted labels are completely lost, i.e. they cannot be recovered and are not converted to incorrect ones.

In this paper, we consider the pool-based active learning with a fixed budget setting, where a finite pool of unlabeled examples is given in advance and we need to sequentially select $N$ examples from the pool to query their labels.
Our setting assumes that abstention feedbacks count towards the budget $N$, so we need to be careful when selecting the queried examples.
Our work takes a Bayesian approach to the problem and learns both the classification model and the unknown abstention rate at the same time.
We call this approach the \emph{Bayesian Active Learning with Abstention Feedbacks} \allowbreak (BALAF) framework.
Our framework can be used to instantiate different algorithms for the active learning with abstention feedbacks problem.

We also contribute to the understandings of this problem both algorithmically and theoretically.
Algorithmically, we develop two novel greedy algorithms for active learning with abstention feedbacks based on our BALAF framework.
Each algorithm uses a different greedy criterion to select queried examples that can give information for both the classification model and the abstention rate.
Theoretically, we prove that our proposed algorithms have theoretical guarantees for a useful utility of the selected examples in comparison to the optimal active learning algorithms.
To the best of our knowledge, these are the first theoretical results for active learning with abstention feedbacks in the Bayesian pool-based setting.

The first greedy algorithm that we propose in this paper aims to maximize the expected version space reduction utility \citep{golovin2011adaptive} of the joint deterministic space deduced from the spaces of possible classification models and abstention rates.
Version space reduction was shown to be a useful utility for active learning \citep{golovin2011adaptive,cuong2013active,cuong2014near}, and our algorithm targets this utility when selecting queried examples.
In essence, the proposed algorithm is similar to the maximum Gibbs error algorithm \citep{cuong2013active} except that we incorporate the terms controlling the estimated abstention rate into the greedy criterion.
By using previous theoretical results for adaptive submodularity \citep{golovin2011adaptive}, we are able to prove that our algorithm has an \emph{average-case near-optimality} guarantee: the average utility value of its selected examples is always within a ${(1-\frac{1}{e})}$ constant factor of the optimal average utility value.

In contrast to the first algorithm, the second algorithm that we propose aims to maximize the worst-case version space reduction utility above.
This algorithm resembles the least confidence active learning algorithm \citep{lewis1994sequential} with the main difference that we also incorporate the estimated abstention rate into the greedy criterion.
From previous theoretical results for pointwise submodularity \citep{cuong2014near}, we can prove that the proposed algorithm has a \emph{worst-case near-optimality} guarantee: the worst-case utility value of its selected examples is always within a ${(1-\frac{1}{e})}$ constant factor of the optimal worst-case utility value.

We conduct experiments to evaluate our proposed algorithms on various binary classification tasks under three different realistic abstention scenarios.
The experiments show that our algorithms are useful compared to the passive learning and normal active learning baselines with various abstention rates under these scenarios.

\section{Related Work}

The theoretical guarantees considered in this paper have been studied for normal Bayesian pool-based active learning where the labeler always gives labels to queried examples \citep{golovin2011adaptive,chen13near,cuong2013active,cuong2014near,chen2015sequential,cuong2016adaptive,cuong2016robustness}.
The theory for the average case was originally developed by \cite{golovin2011adaptive} with adaptive submodular utilities, while that of the worst case was developed by \cite{cuong2014near} for pointwise submodular utilities.
In both cases, $(1-\frac{1}{e})$-factor approximation guarantees were proven for the corresponding greedy algorithms.

The problem of active learning with abstention feedbacks was previously investigated in \cite{fang2012active,ramirez2014anytime,yan2015active}.
\cite{fang2012active} considered a setting similar to ours where the labeler may have knowledge blind spots and would be incapable of labeling examples in such blind spots.
On the other hand, \cite{ramirez2014anytime} studied a situation where the learner may interrupt the labeler rather than waiting for his response, thus allowing the possibility of receiving ``I don't know'' labels.
In both papers, greedy algorithms were used to create a balance between maximizing the received information and minimizing the abstention probability; however, no theoretical guarantees about their performance were obtained.
Our work in this paper, although similar to theirs in spirit, provides theoretical guarantees for the proposed algorithms.

A theoretical work on active learning with abstention feedbacks is \cite{yan2015active}, where both noisy labels and abstention feedbacks are considered.
However, they only examined a simple one-dimensional classification problem and made a low-noise assumption on both the labeling noise and the abstention rate.
Furthermore, the labeler's abstention in their model is non-persistent and that allows the learner to repeatedly query an example until a label is received.
Under this framework, the paper derived an algorithm with a near-optimal asymptotic convergence rate for estimating model parameters.
In contrast, our work in this paper investigates the persistent label scenario, which is much less understood and more difficult to resolve \citep{chen2015sequential,chen2016near}, and we focus on near-optimal query strategies with a finite budget.
Thus, our results are not directly comparable to those in \cite{yan2015active}.

Besides abstention feedbacks, there were other works on active learning with unreliable labelers.
For examples, many authors considered labelers that give incorrect or corrupted labels from various types of noise models \citep{donmez2008proactive,golovin2010near,naghshvar2012noisy,cuong2016robustness,chen2016near}.
\cite{ni2012active} considered a setting where the labeler can return both labels and confidences, while in \cite{malago2014online,zhang2015active}, multiple labelers with different fidelity are available and the learner is given the option of obtaining labels from either weak or strong labelers.
Our work also relates to other works on active learning and adaptive sampling in crowdsourcing such as \cite{yan2011active,zhao2011incremental,mozafari2014scaling,manino2016efficiency,singla2016noisy}.

\section{Pool-Based Active Learning with Abstention Feedbacks}

In pool-based active learning, we are given a finite set (called a pool) $\mathcal{X}$ of unlabeled examples and a budget $N$, and we need to sequentially query the labels of $N$ examples from $\mathcal{X}$ to learn a good classifier.
Normal active learning assumes the human labeler would always give labels for queried examples. 
By contrast, in this paper, we consider active learning with abstention feedbacks where the labeler is allowed to abstain from labeling a queried example.
In other words, the labeler may return ``no label'' to a queried example. 
Our work considers the case where abstention feedbacks count towards the budget $N$, so we need to select queried examples to obtain as many useful labels as possible.

To define the problem, let $\mathcal{Y} = \{ 1, 2, \ldots, \ell \}$ be the set of all possible labels.
Assume there is an unknown true labeling $f_{\text{true}} : \mathcal{X} \rightarrow \mathcal{Y}$ of the whole pool $\mathcal{X}$ that is used by the labeler to label queried examples, and the labeler will return $f_{\text{true}}(x)$ for a queried example $x$ if he decides to label it.
Also assume there is an unknown true abstention pattern $k_{\text{true}} : \mathcal{X} \rightarrow \{ 0, 1\}$ used by the labeler to decide whether or not to label a queried example.
That is, $k_{\text{true}}(x) = 1$ if the labeler abstains from labeling $x$ and $k_{\text{true}}(x) = 0$ if he decides to label it.

\begin{figure*}[t]
\begin{center}
\includegraphics[height=3.4cm]{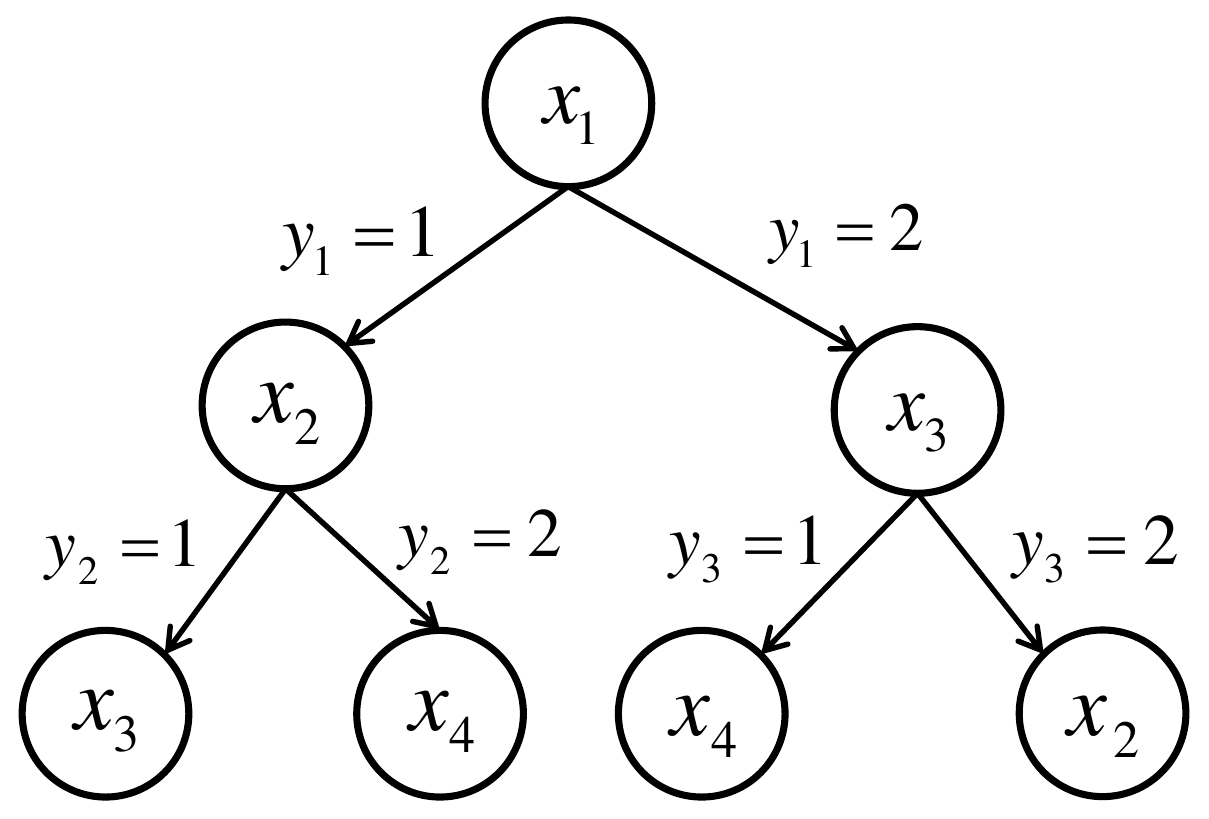}
{\hskip 0.5cm}
\includegraphics[height=3.4cm]{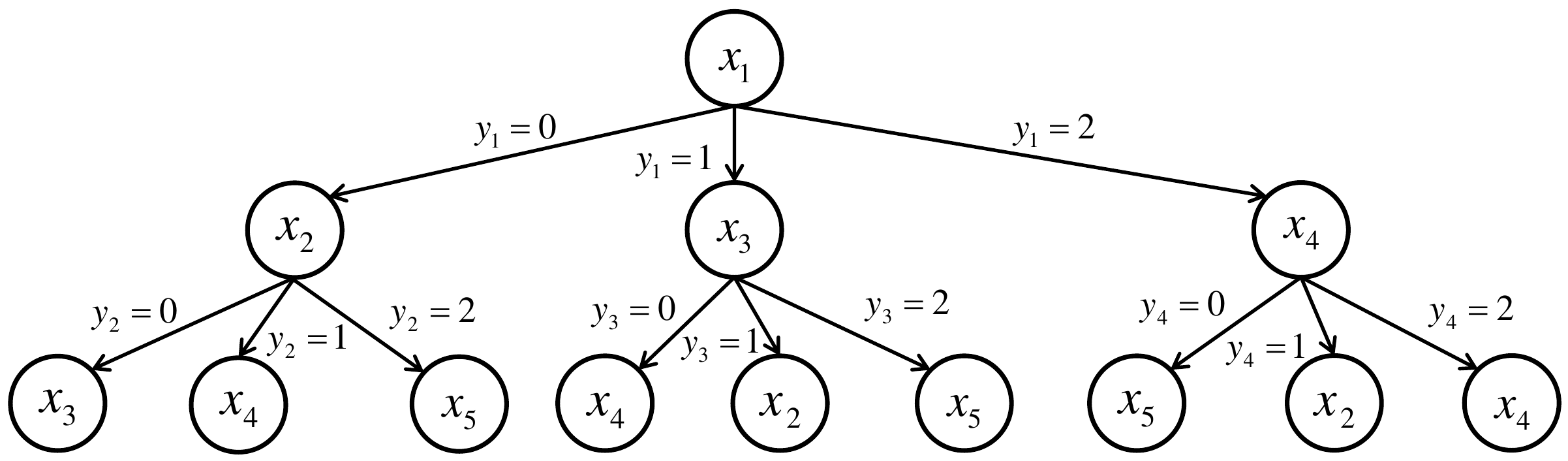}
\caption{Policy trees for pool-based active learning without (upper) and with (bottom) abstention feedbacks. $y_i = 0$ means the labeler abstains from labeling $x_i$ while $y_i = 1$ or $2$ means the labeler gives label $1$ or $2$ for $x_i$ respectively.}
\label{fig:policy-trees}
\end{center}
\end{figure*}

In this setting, an active learning algorithm is a policy for choosing queried examples from $\mathcal{X}$, and these chosen examples depend on the labels as well as abstention feedbacks of previously selected examples.
By definition, a policy is a mapping from a set of examples and the labeler's corresponding responses to the next unlabeled example to be queried, and it can be represented by a policy tree.
During the active learning process, a learner (a policy or algorithm) sequentially selects unlabeled training examples one at a time from $\mathcal{X}$ and asks the labeler for their labels.
The labeler would use $k_{\text{true}}$ to decide whether or not to give the labels, which in turn are determined by $f_{\text{true}}$.
Illustrations of policy trees for normal active learning and active learning with abstention feedbacks are given in Figure \ref{fig:policy-trees}.
Pool-based active learning with abstention feedbacks aims to design algorithms (policies) for selecting queried examples that can give us as much information about $f_{\text{true}}$ (and in some cases, $k_{\text{true}}$) as possible.

\section{Bayesian Active Learning with Abstention Feedbacks (BALAF)}
\label{sec:balaf}

We shall take the Bayesian approach to pool-based active learning with abstention feedbacks, which we call the \emph{Bayesian active learning with abstention feedbacks} (BALAF) framework.
In our framework, we consider a (possibly infinite) set $\mathcal{H}$ of probabilistic hypotheses, where each hypothesis $h \in \mathcal{H}$ is a random function from $\mathcal{X}$ to $\mathcal{Y}$.
Formally, for any $x \in \mathcal{X}$, $h(x)$ is a categorical distribution with probability mass function $\mathbb{P}[h(x)=y]$ for all $y \in \mathcal{Y}$.

Following the Bayesian approach, we assume a prior distribution $p_0[h]$ on $\mathcal{H}$.
If we observe a label $y$ of an example $x$, we can use the following Bayes' rule to obtain a posterior distribution:
\begin{eqnarray*}
p_0[h \mid y,x] &=& \frac{p_0[h] ~ \mathbb{P}[h(x)=y]}{\int p_0[h] ~ \mathbb{P}[h(x)=y] ~ dh} 
\propto p_0[h] ~ \mathbb{P}[h(x)=y]. \\
\end{eqnarray*}
To deal with the abstention feedbacks, we also take the Bayesian approach and consider a set of possible abstention hypotheses $\mathcal{R}$ from $\mathcal{X}$ to $[0,1]$, where each function $r : \mathcal{X} \rightarrow [0,1]$ gives us the \emph{abstention rate} of the examples in $\mathcal{X}$.
More specifically, $r(x)$ is the probability that the labeler abstains from labeling $x$, according to the abstention hypothesis $r$.
We also assume a prior $p_0[r]$ on $\mathcal{R}$.
Note that we have slightly abused the notation $p_0$ for both priors on $\mathcal{H}$ and $\mathcal{R}$.
In this case, $p_0$ can be thought of as a joint distribution on $\mathcal{H} \times \mathcal{R}$ where the two elements are independent, i.e., $p_0[h \wedge r] = p_0[h] ~ p_0[r]$ for $h \in \mathcal{H}$ and $r \in \mathcal{R}$.

During the active learning process, if we receive a label for an example $x$, we can update the posterior distribution using the following Bayes' rule:
\begin{eqnarray*}
p_0[r \mid x \text{ has a label}] &=& \frac{p_0[r] ~ (1-r(x))}{\int p_0[r] ~ (1-r(x)) ~ dr} 
\propto p_0[r] ~ (1-r(x)).
\end{eqnarray*}
Otherwise, if the labeler abstains from labeling $x$, we can update the posterior distribution using:
\begin{eqnarray*}
p_0[r \mid x \text{ has no label}] &=& \frac{p_0[r] ~ r(x)}{\int p_0[r] ~ r(x) ~ dr} 
\propto p_0[r] ~ r(x).
\end{eqnarray*}
We summarize the general BALAF framework in Algorithm \ref{alg:balaf}, where $N$ examples are chosen sequentially and the posteriors are updated according to the above rules.
The framework returns the final posteriors $p_N[h]$ and $p_N[r]$ which can be used to make prediction on new examples or to serve as priors in future active learning processes.
For example, the label distribution of a new example $x$ can be predicted using the posterior $p_N[h]$ by:
\begin{equation*}
p_N[y \mid x] = \int p_N[h] ~ \mathbb{P}[h(x)=y] ~ dh.
\end{equation*}

The posterior $p_N[r]$, on the other hand, can be used as a prior on $r$ in future active learning processes if the same labeler is employed to give labels.
This would enable the learning algorithm to use the prior knowledge about the labeler's preferences to select the most suitable queried examples while avoiding re-learning his abstention patterns from scratch.
The posterior $p_N[r]$ can also be transferred and adapted to other labelers who may have similar abstention patterns.

\begin{algorithm}[t]
\caption{General BALAF framework}
\label{alg:balaf}
\begin{algorithmic}
   \STATE {\bf input:} Priors $p_0[h]$ and $p_0[r]$, budget $N$. \\[3pt]
   \STATE {\bf output:} Final posteriors $p_N[h]$ and $p_N[r]$ after $N$ queries. \\[3pt]
   \FOR{$i=1$ {\bfseries to} $N$}
      \STATE Select unlabeled data point $x^*$ to query based on some active learning criterion. \\[3pt]
      \STATE $y^* \gets$ Query-label($x^*$). \\[3pt]
      \IF{received label $y^*$}
          \STATE Update 
          $p_i[h] \propto p_{i-1}[h] \, \mathbb{P}[h(x^*) = y^*]$, and \\[3pt]
          ${\hskip 1.1cm} p_i[r] \propto p_{i-1}[r] \, (1-r(x^*))$.
      \ELSE
          \STATE Update $p_i[r] \propto p_{i-1}[r] \, r(x^*)$.
      \ENDIF
   \ENDFOR
   \STATE {\bfseries return} $p_N[h]$, $p_N[r]$.
\end{algorithmic}
\end{algorithm}

\section{BALAF Algorithms with Theoretical Guarantees}
\label{sec:two-algo}

In this section, we propose two specific instances of the BALAF framework above that can achieve near-optimality guarantees for reducing the hypothesis spaces that contain $f_{\text{true}}$ and $k_{\text{true}}$.
Our first algorithm provides an average-case near-optimality guarantee, while the second algorithm provides a worst-case near-optimality guarantee.
The algorithms only differ in the ways we choose the queried data point $x^*$ in Algorithm \ref{alg:balaf}.

In what follows, for any ${ S = \{ x_1, x_2, \ldots, x_n \} \subseteq \mathcal{X} }$ and $h \in \mathcal{H}$, we define ${ h(S) \triangleq \{ h(x_1), h(x_2), \ldots, h(x_n) \} }$.
We shall assume $h(x_i)$ and $h(x_j)$ are independent for any fixed $h$ and $i \neq j$.
Thus, $h(S)$ is also a categorical distribution with probability mass function $\mathbb{P}[h(S)=\mathbf{y}] = \prod_{i = 1}^n \mathbb{P}[h(x_i)=y_i]$ for all ${ \mathbf{y} = \{ y_1, y_2, \ldots, y_n \} \in \mathcal{Y}^{|S|} }$.
We call $\mathbf{y}$ a labeling of $S$ as it contains the labels of the examples in $S$.

For any $S \subseteq \mathcal{X}$ and any distribution $p[h]$ on $\mathcal{H}$, let $\mathbf{Y}$ be the random variable for the labeling of $S$ with respect to the distribution $p$.
We note that $\mathbf{Y}$ takes values in $\mathcal{Y}^{|S|}$ with probability mass function: 

\begin{equation}
\label{eq:marginal}
p[\mathbf{Y} = \mathbf{y};S] = \int p[h] ~ \mathbb{P}[h(S) = \mathbf{y}] ~ dh
\end{equation}
for all $\mathbf{y} \in \mathcal{Y}^{|S|}$.
This is also the marginal probability that the labeling of $S$ is $\mathbf{y}$.
As a special case, if $S$ is a singleton $\{ x \}$ and $Y$ is the random variable for the label of $x$, we write $p[Y=y;x]$ for $y \in \mathcal{Y}$ to denote the probability mass function of $Y$.

\begin{figure*}[t]
\begin{center}
\includegraphics[height=4.3cm]{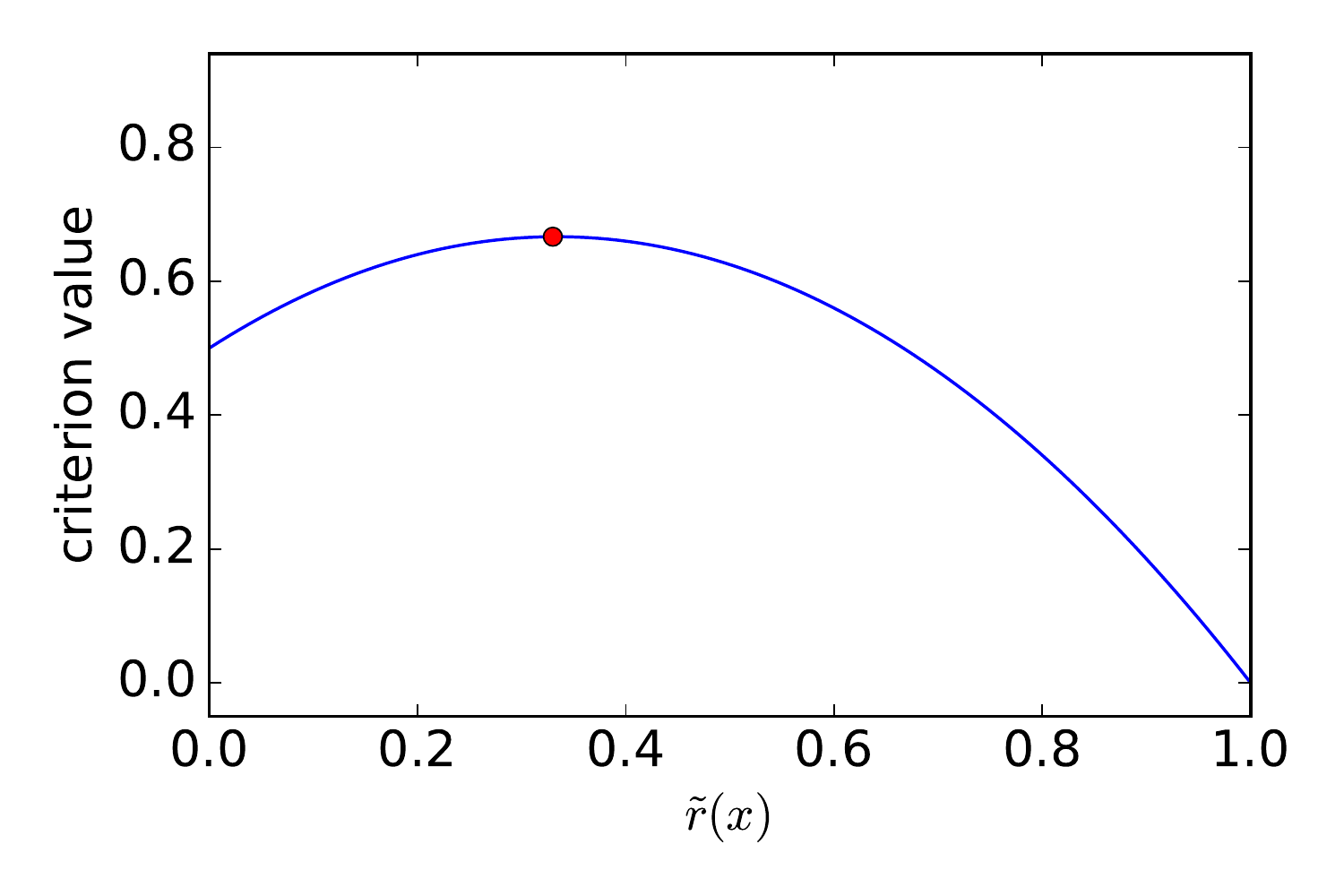}
{\hskip 5mm}
\includegraphics[height=4.3cm]{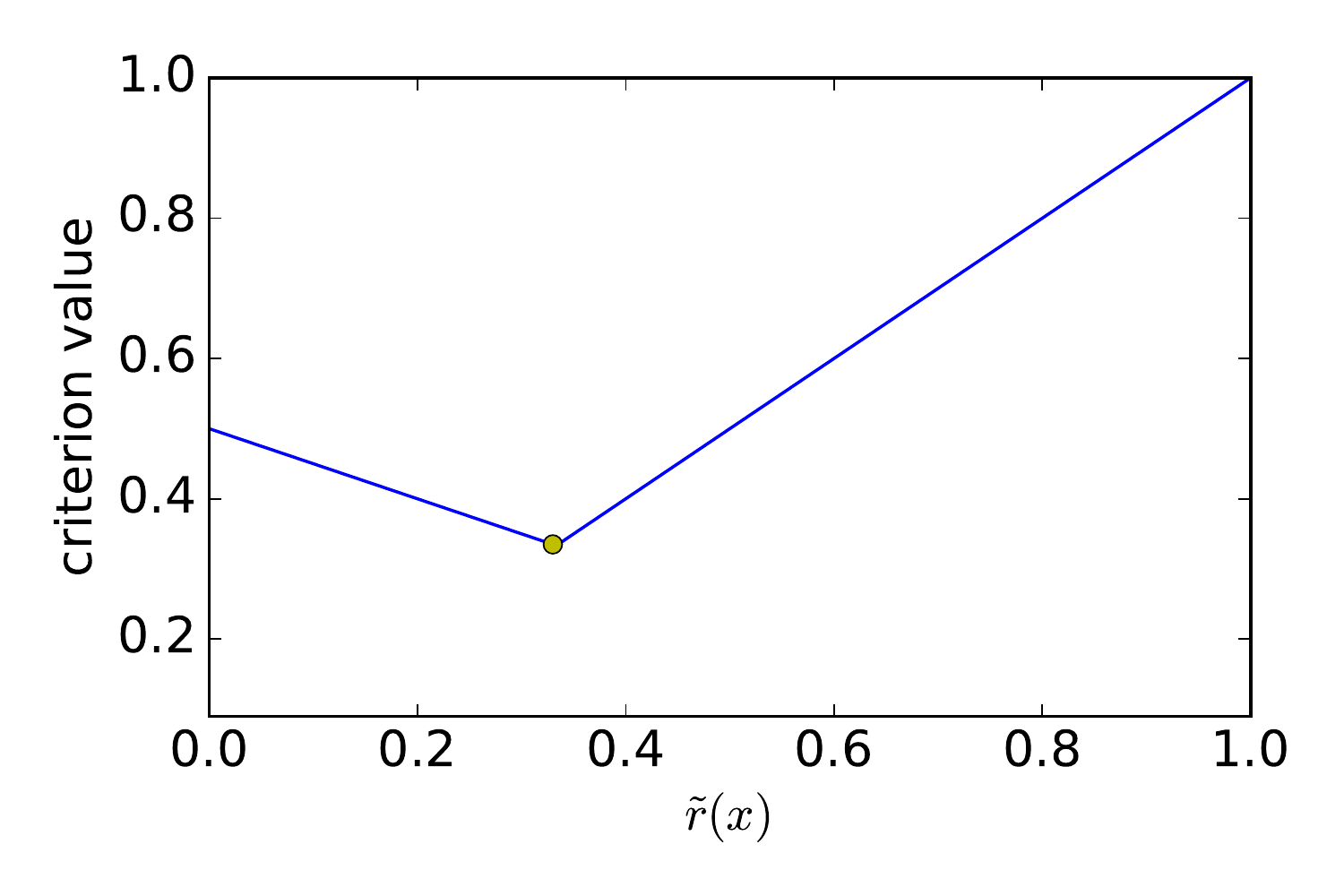}
\caption{Graphs showing the greedy criterion values in the average-case BALAF (upper) and worst-case BALAF (bottom) algorithms as a function of $\tilde{r}(x)$ in a binary classification problem. The graphs are plotted with the fixed distribution $p_{i-1}[Y=1;x] = p_{i-1}[Y=2;x] = 0.5$. The red and yellow points indicate the maximum and minimum points in the graphs respectively.}
\label{fig:criteria}
\end{center}
\end{figure*}

\subsection{The Average-case BALAF Algorithm}

In this average-case BALAF algorithm, at each iteration $i$ in Algorithm \ref{alg:balaf}, we select the queried data point $x^*$ as follows.
First, we estimate the average abstention function $\tilde{r}(x)$ based on the current posterior $p_{i-1}[r]$:

\begin{equation}
\tilde{r}(x) \triangleq \mathbb{E}_{r \sim p_{i-1}}[r(x)] = \int p_{i-1}[r] ~ r(x) ~ dr.
\end{equation}

Then we select the example $x^*$ to query using the following greedy criterion:
\begin{equation}
\label{eq:criterion1}
x^* = \arg \max_{x \in \mathcal{X}} \{ 1 - \tilde{r}(x)^2 - (1 - \tilde{r}(x))^2 \sum_{y \in \mathcal{Y}} p_{i-1}[Y = y; x]^2 \},
\end{equation}

Intuitively, this criterion maximizes the expected one-step utility increment, with the utility function being defined in Equation \eqref{eq:utility} below.
Equation \eqref{eq:criterion1} resembles the maximum Gibbs error criterion \citep{cuong2013active} which selects
\[ x^* = \arg \max_x \{ 1 - \sum_y p_{i-1}[Y=y;x]^2 \}, \]
except that we incorporate the terms $\tilde{r}(x)^2$ and ${ (1 - \tilde{r}(x))^2 }$ into the criterion.

If we fix the distribution $p_{i-1}[Y=y;x]$, the criterion value in Equation \eqref{eq:criterion1} achieves its maximum when ${ \tilde{r}(x) = \sum_y p_{i-1}[Y=y;x]^2/(1 + \sum_y p_{i-1}[Y=y;x]^2) }$ and achieves its minimum when $\tilde{r}(x) = 1$.
Figure \ref{fig:criteria} (left) illustrates this criterion value as a function of $\tilde{r}(x)$.
Thus, given $\tilde{r}(x)$ reasonably approximates the true abstention rate, this algorithm would give more preference to the examples with abstention rate near $\sum_y p_{i-1}[Y=y;x]^2/(1 + \sum_y p_{i-1}[Y=y;x]^2)$ and less preference to the examples with abstention rate near $1$.

\textbf{Near-optimality Guarantee}.
We now show the average-case near-optimality guarantee for this algorithm.
In the context of this paper, near-optimality means the algorithm can achieve a constant factor approximation to the optimal algorithm with respect to some objective function.

To define an objective function that is useful for active learning with abstention feedbacks, we first induce a deterministic hypothesis space equivalent to the original probabilistic hypothesis space $\mathcal{H}$.
In particular, consider the hypothesis space ${ \mathcal{F} \triangleq \{ f : \mathcal{X} \rightarrow \mathcal{Y} \} }$ consisting of all deterministic functions from $\mathcal{X}$ to $\mathcal{Y}$.
We induce a new prior $q_0$ on $\mathcal{F}$ from the original prior $p_0$ such that ${ q_0[f] \triangleq p_0[\mathbf{Y} = f(\mathcal{X});\mathcal{X}] }$, the marginal probability that the labeling of the whole pool $\mathcal{X}$ is $f(\mathcal{X})$.
For any $S \subseteq \mathcal{X}$ and ${ \mathbf{y} \in \mathcal{Y}^{|S|} }$, we can define $q_0[\mathbf{Y} = \mathbf{y};S]$ similarly to Equation \eqref{eq:marginal} with the hypothesis space $\mathcal{F}$ and distribution $q_0$.

Also consider the space $\mathcal{K} \triangleq \{ k : \mathcal{X} \rightarrow \{0,1\} \}$ consisting of all deterministic functions from $\mathcal{X}$ to $\{0,1\}$.
In essence, $k(x) = 1$ means the labeler abstains from labeling $x$ while $k(x) = 0$ means the labeler gives a label for $x$.
We will call each $k \in \mathcal{K}$ an \emph{abstention pattern}.
The prior $p_0[r]$ also induces a probability distribution $q_0[k]$ on $\mathcal{K}$ where:

\begin{eqnarray*}
q_0[k] &\triangleq& \int p_0[r] ~ P_r[k] ~ dr, \text{ and} \\
P_r[k] &\triangleq& \prod_{x \in \mathcal{X}} (1-r(x))^{1-k(x)} r(x)^{k(x)} \\
\end{eqnarray*}
is the probability (with respect to the rate $r$) that the labeler gives or abstains from giving labels to the whole pool $\mathcal{X}$ according to the abstention pattern $k$.
For any ${ S \subseteq \mathcal{X} }$ and $\mathbf{z} \in \{ 0,1 \}^{|S|}$, we can define $q_0[\mathbf{Z} = \mathbf{z};S]$ similarly to Equation \eqref{eq:marginal} with the hypothesis space $\mathcal{K}$ and distribution $q_0$, where $\mathbf{Z}$ is the random variable for the abstention pattern of $S$.
Note that the induced prior $q_0$ can also be thought of as a joint prior on $\mathcal{F} \times \mathcal{K}$ where the two elements are independent, i.e., $q_0[f \wedge k] = q_0[f] ~ q_0[k]$.

For $S \subseteq \mathcal{X}$, $f \in \mathcal{F}$, and $k \in \mathcal{K}$, we consider the utility function:
\begin{equation}
\label{eq:utility}
g(S, (f,k)) \triangleq 1 - q_0[\mathbf{Y} = f(S) \wedge \mathbf{Z} = k(S);S],
\end{equation}
where ${q_0[\mathbf{Y} = f(S) \wedge \mathbf{Z} = k(S);S] \triangleq q_0[\mathbf{Y} = f(S);S]} \times \allowbreak {q_0[\mathbf{Z} = k(S);S]}$
is the joint marginal probability (with respect to $q_0$) that the labeling of $S$ is $f(S)$ and the abstention pattern of $S$ is $k(S)$.
This is a useful utility function for active learning because it is the version space reduction utility with respect to the joint prior $q_0[f \wedge k]$ on the joint space $\mathcal{F} \times \mathcal{K}$ \citep{golovin2011adaptive}.

With this utility, our objective function is defined as:
\begin{equation}
G_{\text{avg}}(\pi) \triangleq \mathbb{E}_{f_{\text{true}},k_{\text{true}} \sim q_0} [g(\mathbf{x}^{\pi}_{f_{\text{true}},k_{\text{true}}}, (f_{\text{true}}, k_{\text{true}}))],
\end{equation}
where for all $f$ and $k$, $\mathbf{x}^{\pi}_{f,k}$ is the set of examples selected by the policy $\pi$ given that the true labeling is $f$ and the true abstention pattern is $k$.
This objective function is the average of the above utility with respect to the joint prior $q_0[f \wedge k]$.
Note that in this objective, $f_{\text{true}}$ and $k_{\text{true}}$ are drawn from the prior since we operate in the Bayesian setting.
The following theorem proves the average-case near-optimality guarantee for the average-case BALAF algorithm.
The proof of this theorem is given in the Appendix.

\begin{theorem}
\label{theorem:near-opt-avg}
For any budget $N \ge 1$, let $\pi$ be the policy selecting $N$ examples using the average-case BALAF algorithm and let $\pi^*_{\text{avg}}$ be the optimal policy with respect to $G_{\text{avg}}$ that selects $N$ examples. We have:
\[ G_{\text{avg}}(\pi) > (1 - \frac{1}{e}) ~ G_{\text{avg}}(\pi^*_{\text{avg}}). \]
\end{theorem}

\subsection{The Worst-case BALAF Algorithm}

The worst-case BALAF algorithm is essentially similar to the previous average-case BALAF algorithm, except that we replace the greedy criterion in Equation \eqref{eq:criterion1} by the following greedy criterion:
\begin{equation}
\label{eq:criterion2}
x^* = \arg \min_{x \in \mathcal{X}} \{ \max \{ \tilde{r}(x), (1-\tilde{r}(x)) \max_{y \in \mathcal{Y}} p_{i-1}[Y = y; x]  \} \}.
\end{equation}
Intuitively, this criterion maximizes the worst-case one-step utility increment, with the version space reduction utility in Equation \eqref{eq:utility}.
The criterion \eqref{eq:criterion2} resembles the least confidence criterion \citep{lewis1994sequential}, which selects 
\[ x^* = \arg \min_x \{ \max_y p_{i-1}[Y=y;x] \}, \] 
except that we also incorporate the terms $\tilde{r}(x)$ and $1 - \tilde{r}(x)$ into the criterion.

If we fix the distribution $p_{i-1}[Y=y;x]$, the criterion value in Equation \eqref{eq:criterion2} achieves its maximum when $\tilde{r}(x) = 1$ and achieves its minimum when $\tilde{r}(x) =$ \allowbreak $\max_y p_{i-1}[Y = y; x]  \}/(1 + \max_y p_{i-1}[Y = y; x]) $.
Figure \ref{fig:criteria} (right) illustrates this criterion value as a function of $\tilde{r}(x)$.
Thus, given $\tilde{r}(x)$ reasonably approximates the true abstention rate, this algorithm would give more preference to the examples with abstention rate near $\max_y p_{i-1}[Y = y; x]  \}/(1 + \max_y p_{i-1}[Y = y; x])$ and less preference to the examples with abstention rate near $1$.

\textbf{Near-optimality Guarantee}.
We now show the worst-case near-optimality guarantee for this algorithm.
For this guarantee, we still make use of the version space reduction utility function $g(S, (f,k))$ defined in Equation \eqref{eq:utility}.
Using this utility, we define the following worst-case objective function:
\begin{equation}
G_{\text{worst}}(\pi) \triangleq \min_{(f,k) \in \mathcal{F} \times \mathcal{K}} [g(\mathbf{x}^{\pi}_{f,k}, (f,k))].
\end{equation}

This objective function is the worst possible utility achieved by the policy $\pi$.
The following theorem proves the worst-case near-optimality guarantee for this algorithm.
The proof of this theorem is given in the Appendix.

\begin{theorem}
\label{theorem:near-opt-worst}
For any budget $N \ge 1$, let $\pi$ be the policy selecting $N$ examples using the worst-case BALAF algorithm and let $\pi^*_{\text{worst}}$ be the optimal policy with respect to $G_{\text{worst}}$ that selects $N$ examples. We have:
\[ G_{\text{worst}}(\pi) > (1 - \frac{1}{e}) ~ G_{\text{worst}}(\pi^*_{\text{worst}}). \]
\end{theorem}

\section{Experiments}

In this section, we experimentally evaluate the proposed BALAF algorithms.
In particular, we compare four algorithms:
\begin{itemize}
\item PL: the passive learning baseline with randomly selected examples,
\item ALg: the active learning baseline using the maximum Gibbs error criterion \citep{cuong2013active},
\item ALa: the average-case BALAF algorithm, and
\item ALw: the worst-case BALAF algorithm.
\end{itemize}

For binary classification, ALg is equivalent to other well-known active learning algorithms such as the least confidence \citep{lewis1994sequential} and maximum entropy \citep{settles2010active} algorithms.
The PL and ALg baselines do not learn the abstention probability of the examples, i.e., they ignore whether an example would be labeled or not when making a decision.
In contrast, the proposed algorithms ALa and ALw take into account the estimated abstention probability $\tilde{r}(x)$ when making decisions.

\begin{figure}[t]
\begin{center}
\includegraphics[width=0.54\textwidth]{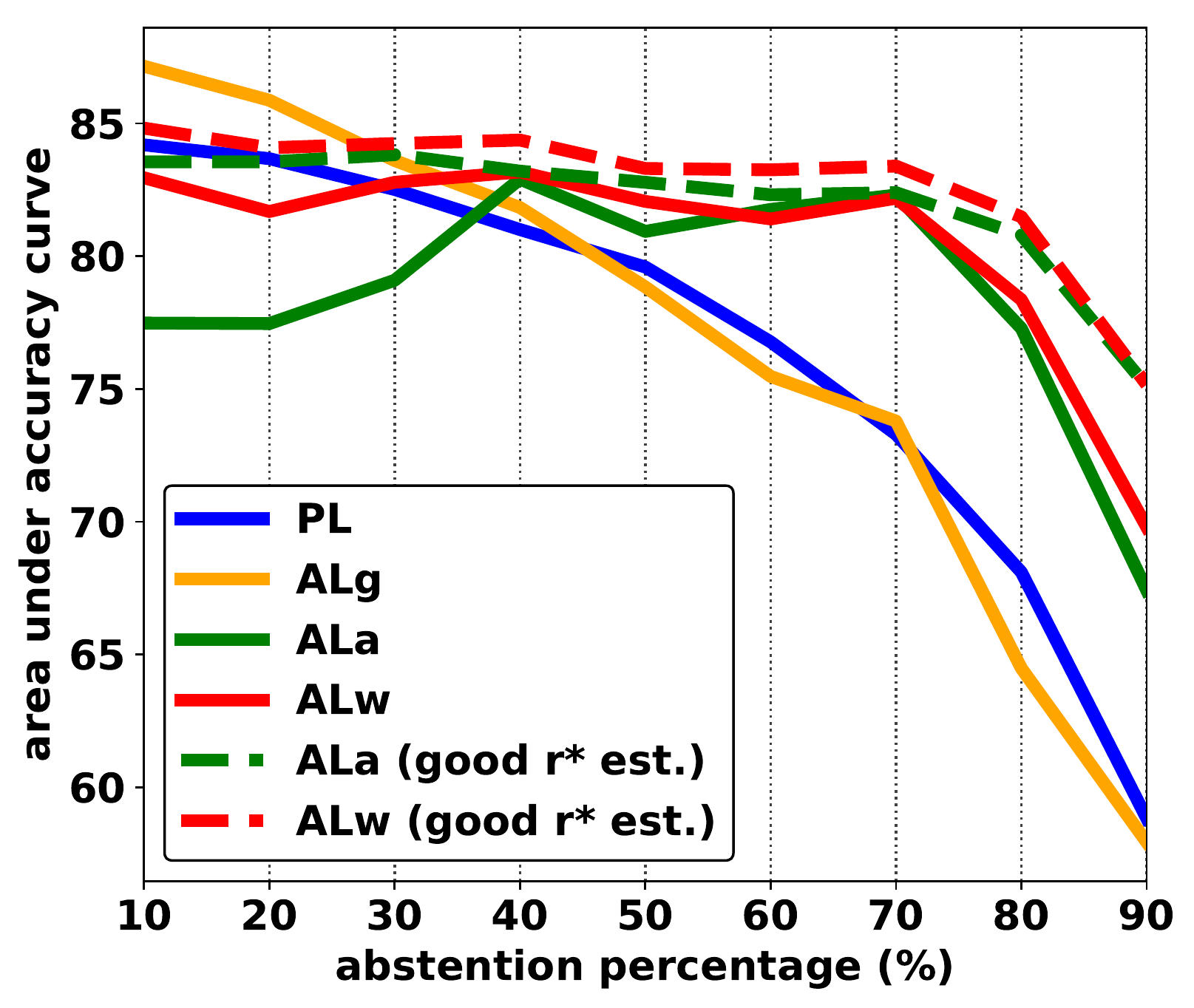}
\caption{AUAC scores with labeler abstaining on examples unrelated to target task.}
\label{fig:set_unrelate}
\end{center}
\end{figure}

To show the potential of our algorithms further, we also consider two variants of ALa and ALw that are assumed to know a good estimate of the training examples' abstention rates $r^*(x)$.
In particular, for these versions of ALa and ALw (shown as dashed lines in Figures \ref{fig:set_unrelate} and \ref{fig:set_easy_hard}), we train a logistic regression model using the actual abstention pattern on the whole training set to predict the abstention probability for each example.
We keep this classifier fixed throughout the experiments and use it to estimate $\tilde{r}(x)$ in these versions of ALa and ALw.

In the algorithms, we use Bayesian logistic regression models for both $\mathcal{H}$ and $\mathcal{R}$.
That is, each hypothesis $h \in \mathcal{H}$ and each abstention hypothesis $r \in \mathcal{R}$ is a logistic regression model.
We put an independent Gaussian prior $\mathcal{N}(0,\sigma^2)$ on each parameter of the logistic regression models (for both $\mathcal{H}$ and $\mathcal{R}$).
In this case, the posteriors are proportional to the regularized likelihood of the observed data with $\ell_2$ penalty.
Since we experiment with data sets containing very high dimensional data (more than 61,000 dimensions), running MCMC or even variational inference is very slow.
Thus, for efficiency, we use the maximum a posteriori (MAP) hypotheses to estimate the probabilities in our algorithms.
Finding the MAP hypotheses is equivalent to maximizing the regularized log-likelihood of the observed data.

Following previous works in active learning \citep{settles2008analysis,cuong2013active,cuong2014near}, we evaluate the algorithms using the area under the accuracy curve (AUAC) scores.
For each task in our experiments, we compute the scores on a separate test set during the first 300 queries and then normalize these scores so that their values are between 0 and 100.
The final scores are obtained by averaging 10 runs of the algorithms using different random seeds.

We shall consider three scenarios: (1) the labeler abstains from labeling examples unrelated to the target classification task, (2) the labeler abstains from labeling easy examples, and (3) the labeler abstains from labeling hard examples.

\subsection{Abstention on Data Unrelated to Target Task}

We consider the binary text classification task between two recreational topics: \emph{rec.motorcycles} and \emph{rec.sport.baseball} from the 20 Newsgroups data \citep{joachims1996probabilistic}.
In the pool of unlabeled data, we allow examples from other classes (e.g., in the \emph{computer} category) that are not related to the two target classes.
The labeler always abstains from labeling these redundant examples while always giving labels for examples from the target classes.
Thus, the abstention is on examples unrelated to the target task, and this satisfies the independence assumption between $h$ and $r$ (or between $f$ and $k$) in Sections \ref{sec:balaf} and \ref{sec:two-algo}.
In the experiment, we fix the pool size to be 1322 and vary the abstention percentage (\%) of the labeler by changing the ratio of the redundant examples.

Figure \ref{fig:set_unrelate} shows the results for various abstention percentages.
From the figure, our algorithms ALa and ALw are consistently better than the baselines for abstention percentages above 40\%.
When a good estimate of $r^*$ is available, our algorithms perform better than all the other algorithms for abstention percentages above 30\%.
This shows the advantage of modeling the labeler's abstention pattern in this setting, especially for medium to high abstention percentages.

\subsection{Abstention on Easy Examples}

\begin{figure*}[t]
\begin{center}
\includegraphics[width=0.24\textwidth]{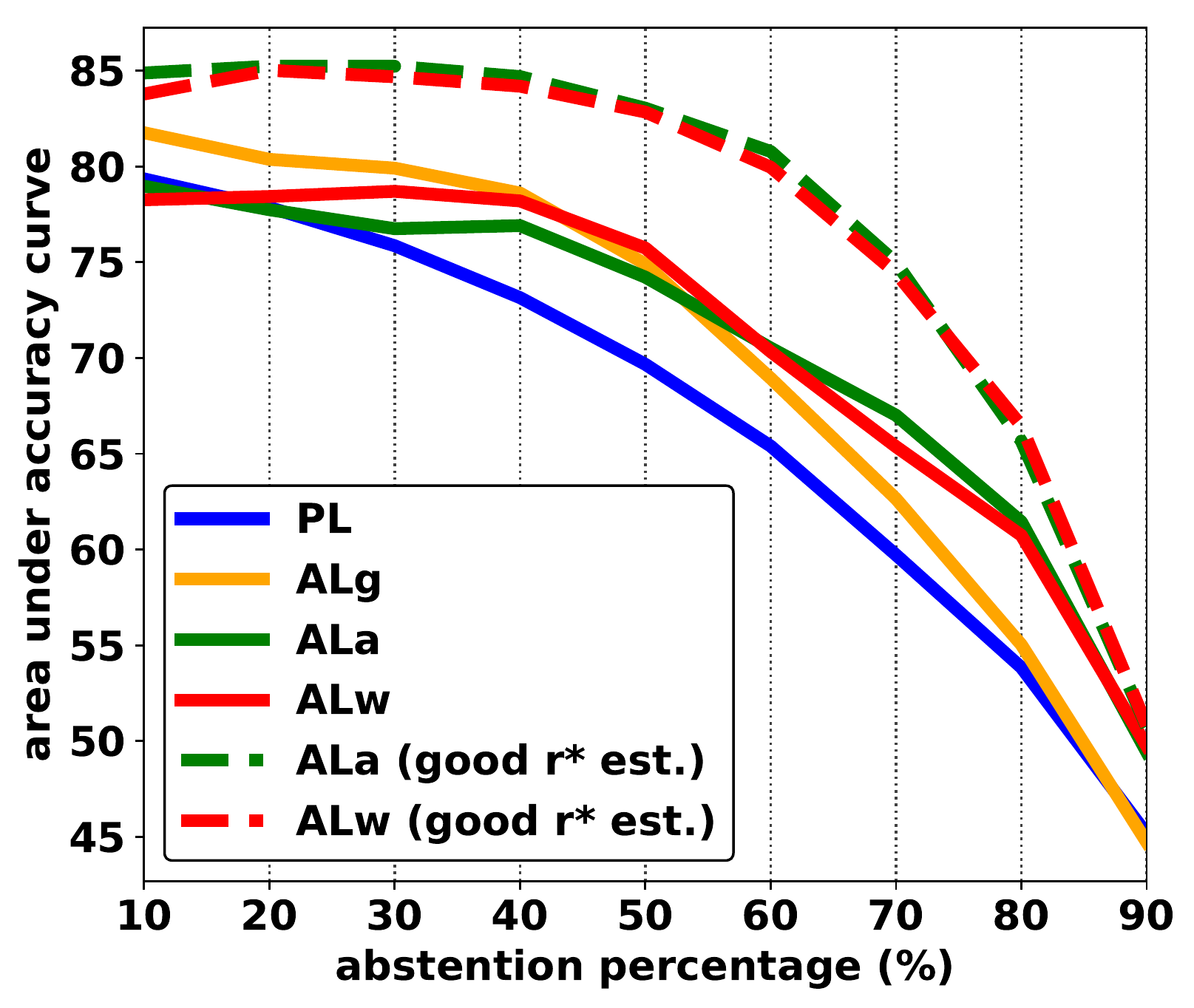}
\includegraphics[width=0.24\textwidth]{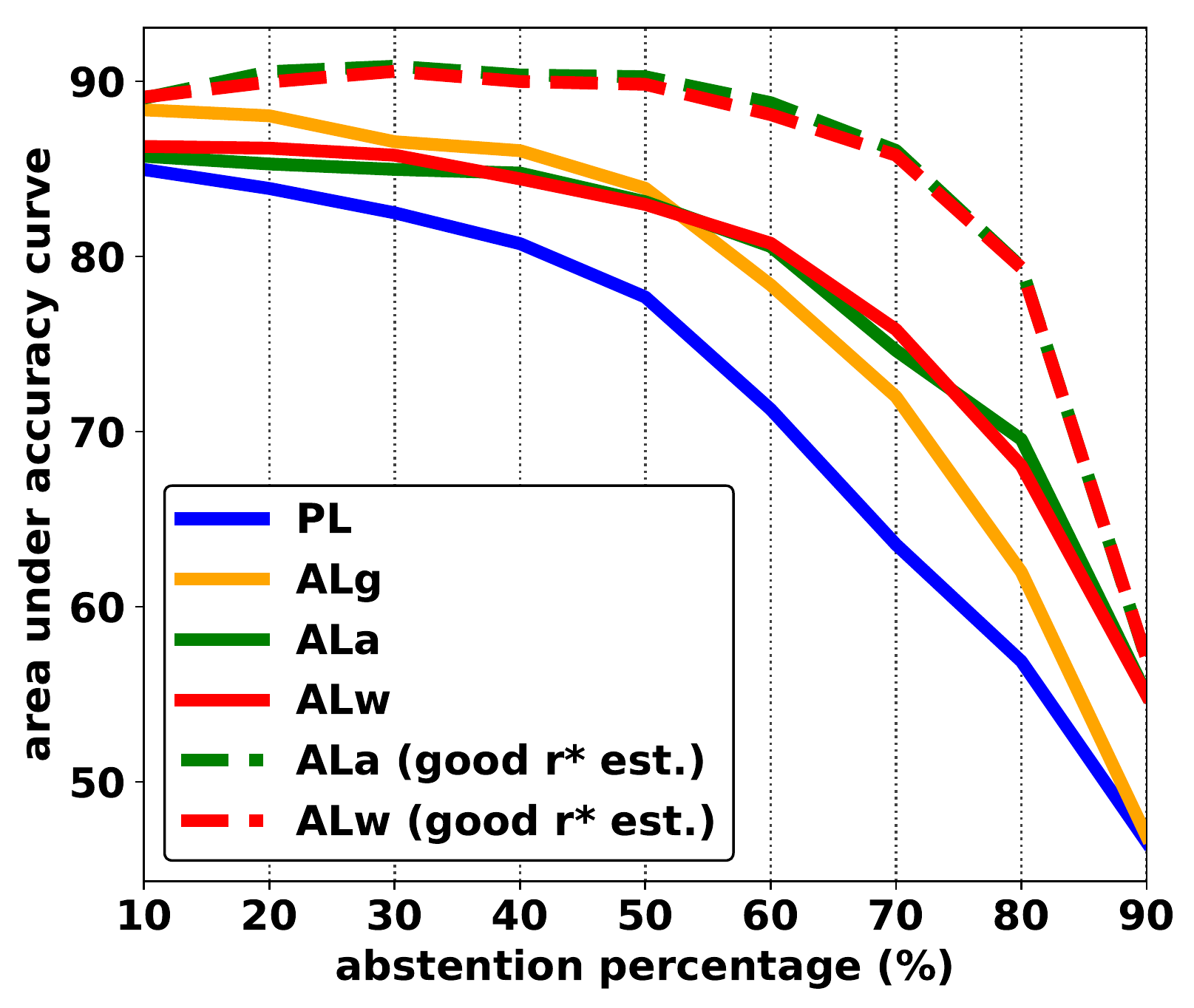}
\includegraphics[width=0.24\textwidth]{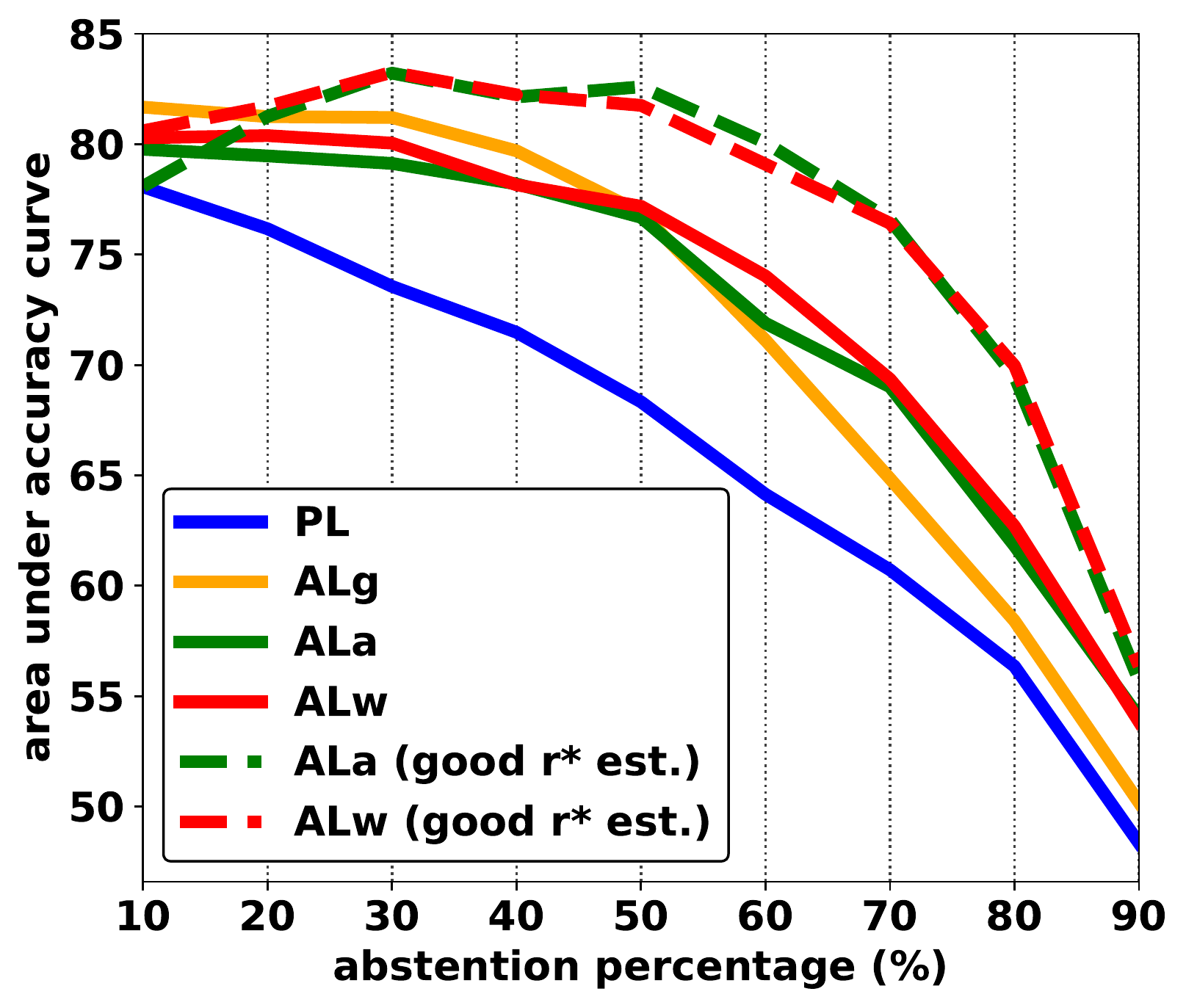}
\includegraphics[width=0.24\textwidth]{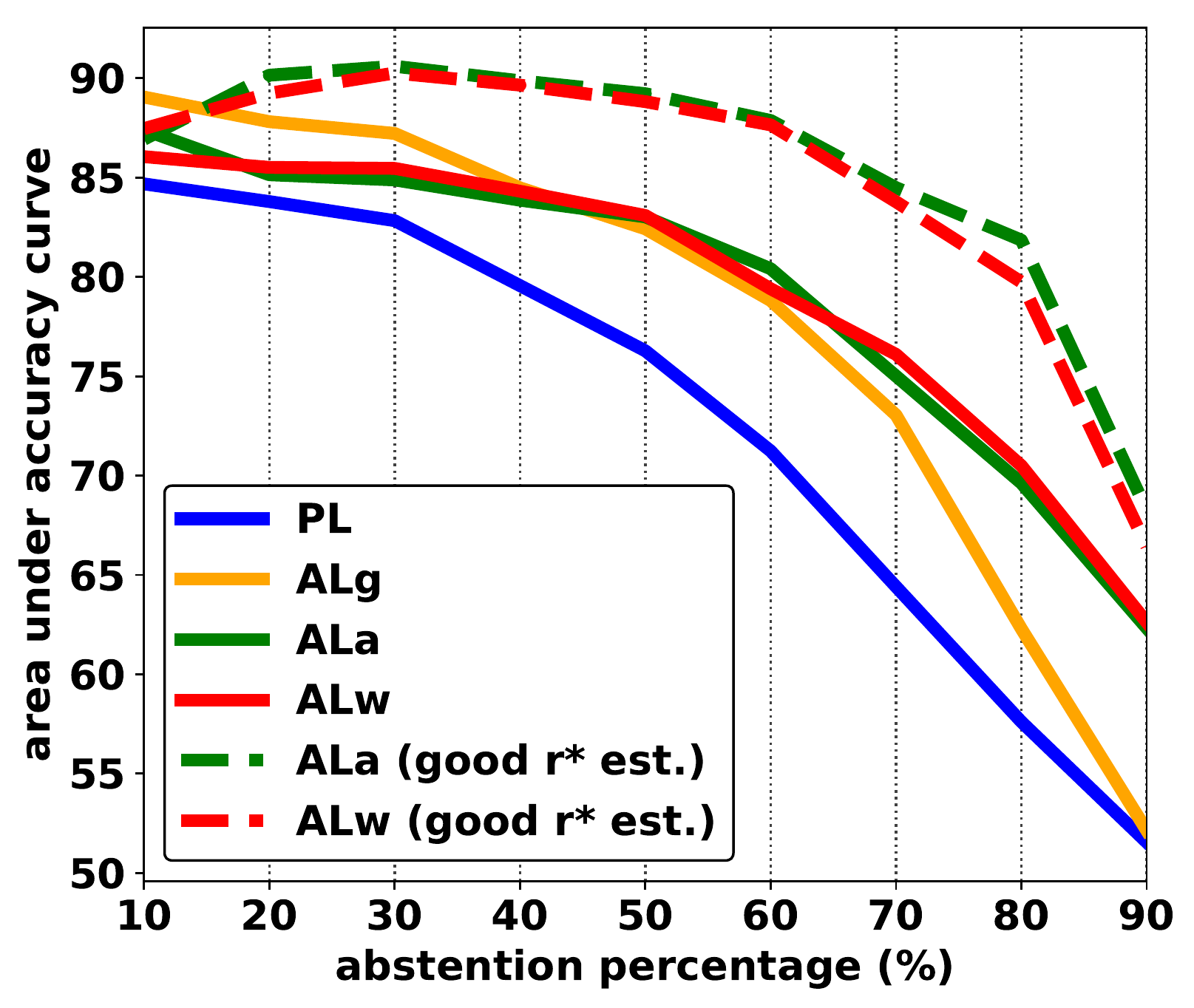}
\includegraphics[width=0.24\textwidth]{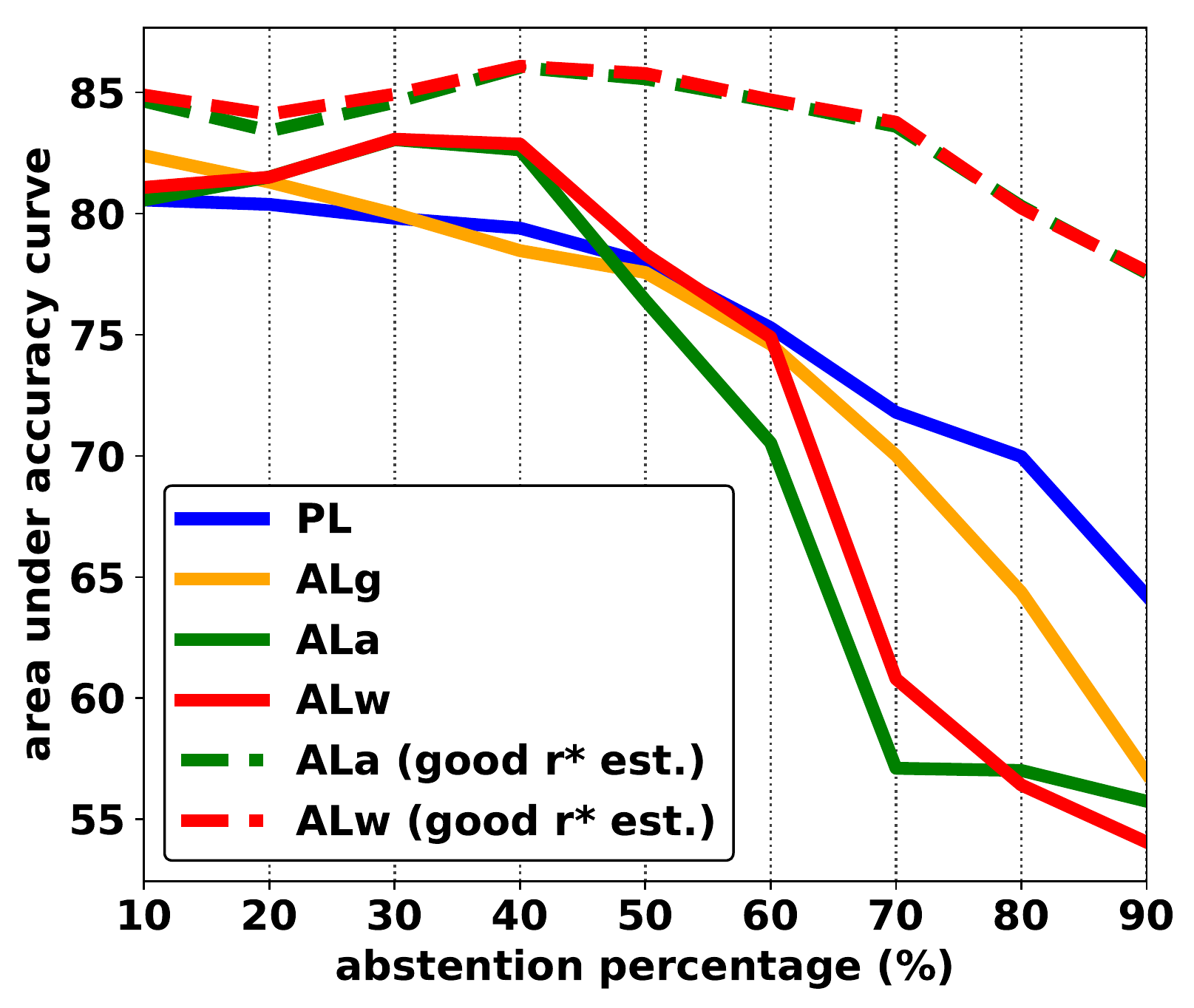}
\includegraphics[width=0.24\textwidth]{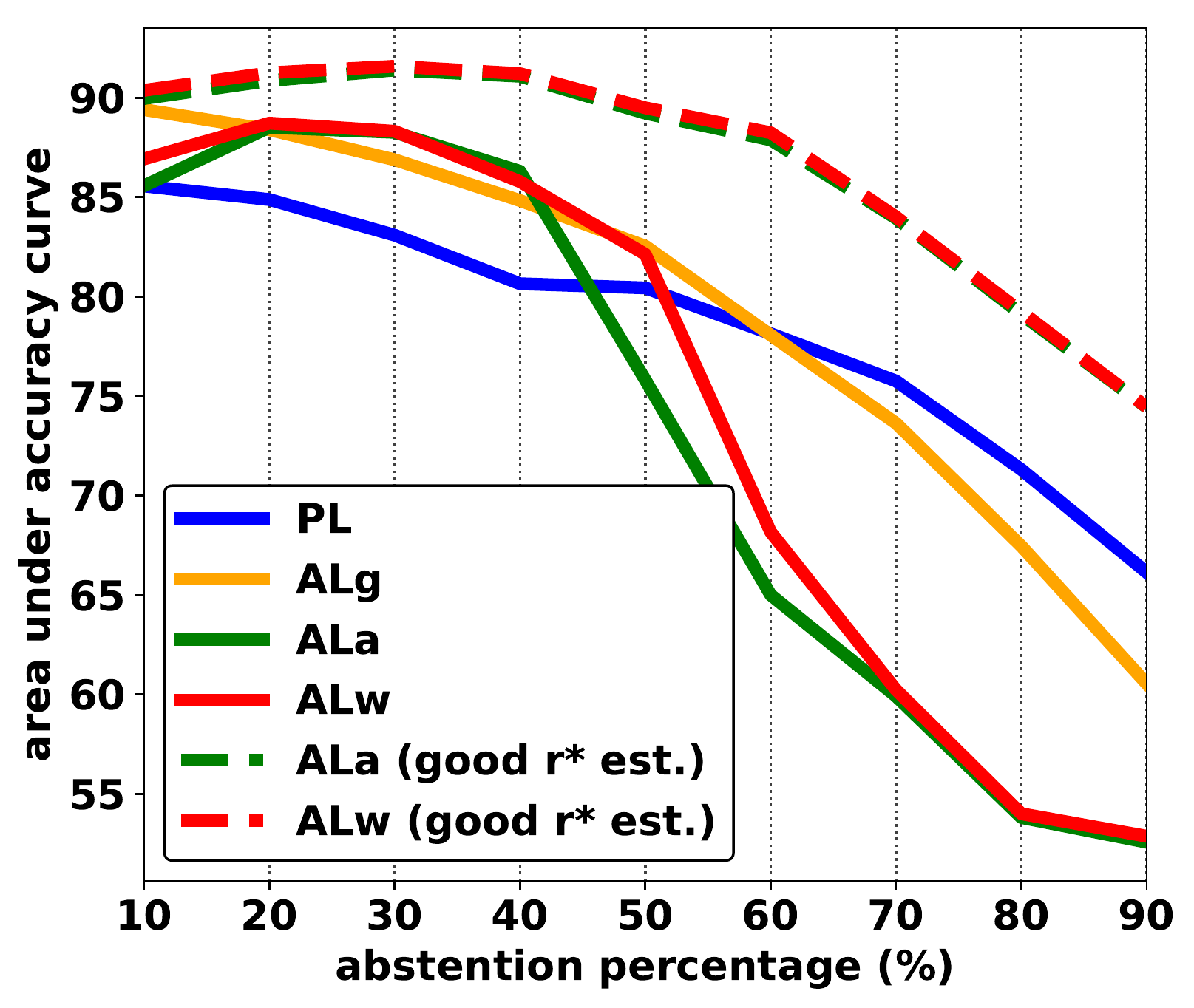}
\includegraphics[width=0.24\textwidth]{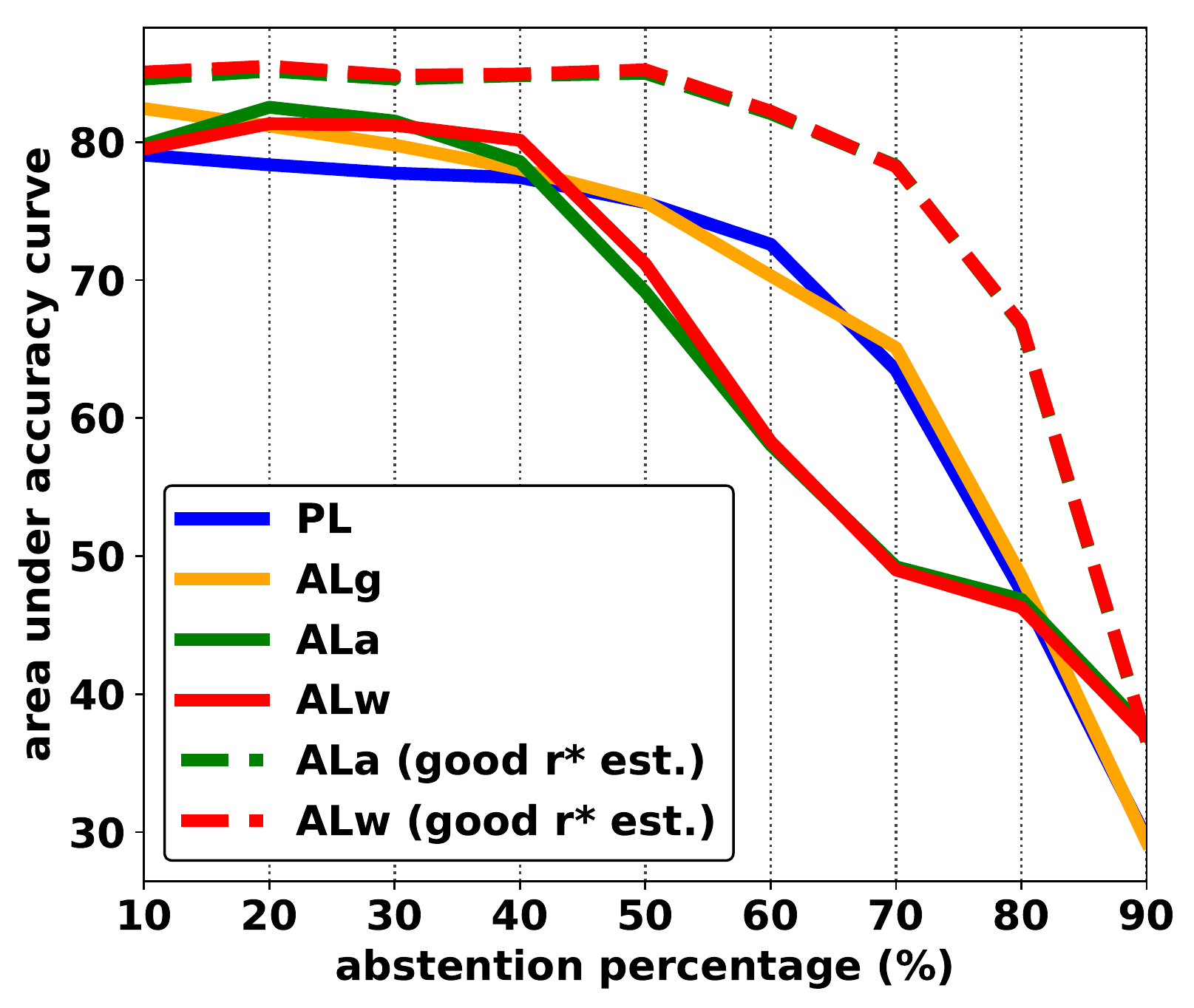}
\includegraphics[width=0.24\textwidth]{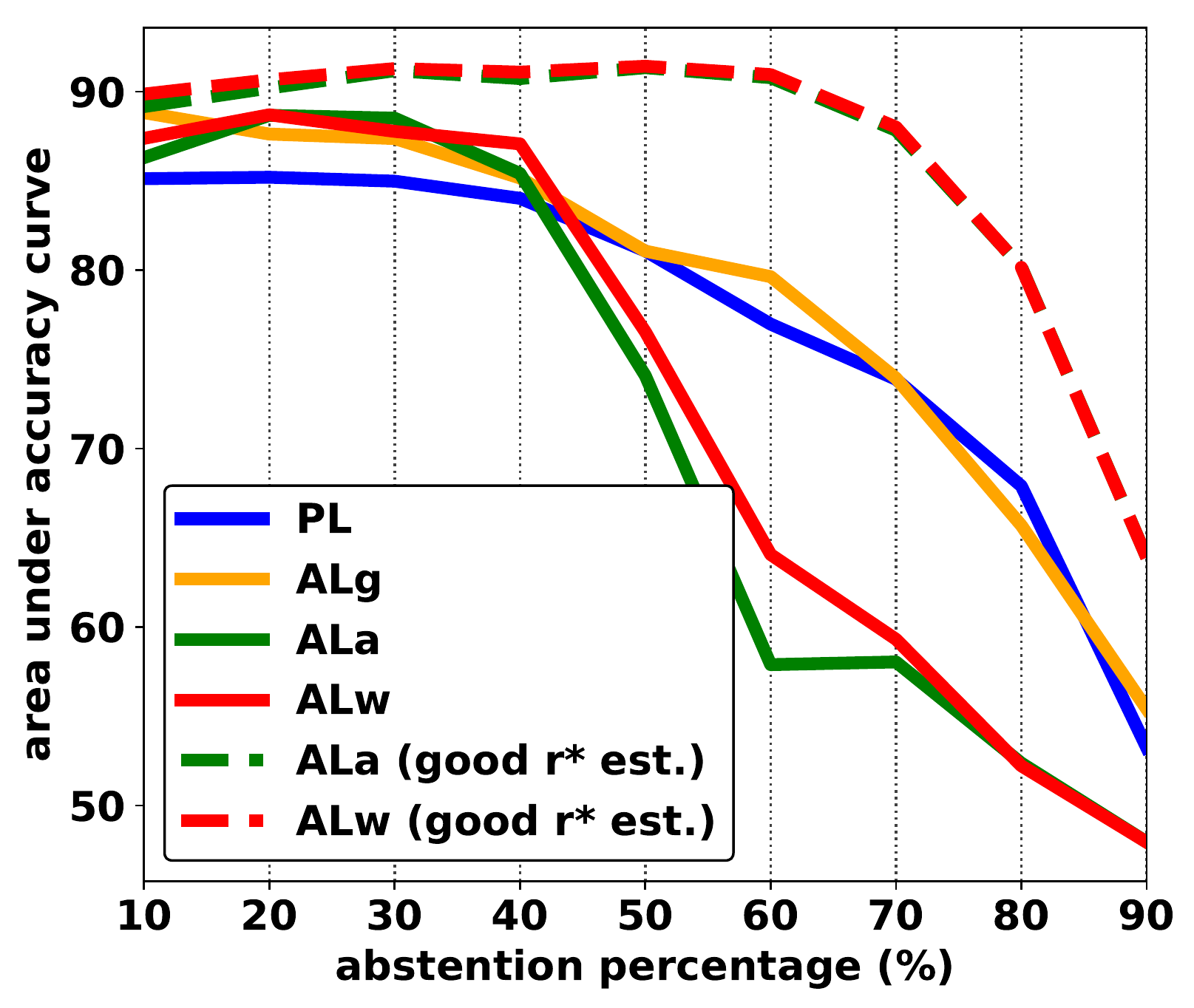}
\caption{AUAC scores with a labeler abstaining on easy examples (first row) and on hard examples (second row).}
\label{fig:set_easy_hard}
\end{center}
\end{figure*}

In this scenario, we test with the labeler who abstains from labeling easy data, which are far from the true decision boundary.
This setting may seem counter-intuitive, but it is in fact not unrealistic.
For example in the learning with corrupted labels setting discussed in Section \ref{sec:intro}, easy examples may be considered less important than hard examples and thus were less protected than hard ones.
In this case, an attacker may attempt to corrupt the labels of those easy examples to bring down the performance of the learned classifier.
Furthermore, under a heavy attack, we may expect a high abstention percentage.
As another example, in medical diagnosis, lung cancer screening is only recommended for the high-risk group (heavy smoking, 55-74 years old, etc.) \citep{roberts2013screening}, so labels (cancer or no cancer) for the low-risk group (easy data) are often unavailable.

We simulate the abstention pattern for this scenario by first learning a logistic regression model with regularizer $\sigma^2 = 0.5$ on the whole training data set and then measuring the distance between the model's prediction probability to 0.5 for each example.
The labeler would always abstain from labeling the subset of the training data (with size depending on the abstention percentage) that have the largest such distances while he would always give labels for the other examples.
Figure \ref{fig:set_easy_hard} (first row) shows the results for this setting on the following 4 binary text classification data sets from the 20 Newsgroups data (from left to right):
\begin{itemize}
\item comp.sys.mac.hardware / comp.windows.x, 
\item rec.motorcycles / rec.sport.baseball, 
\item sci.crypt / sci.electronics, 
\item sci.space / soc.religion.christian.
\end{itemize}

From the results, ALa and ALw work very well when the abstention percentage is above 50\%.
This shows that it is useful to learn and take into account the abstention probabilities when the abstention percentage is high (e.g., under heavy attacks), and our algorithms provide a good way to exploit this information.
When the abstention percentage is small, the advantages of ALa and ALw diminish.
This is expected because in this scenario, learning the abstention pattern is more expensive than simply ignoring it.
However, when a good estimate of $r^*$ is available, ALa and ALw perform better than all the other algorithms for most abstention percentages.

\subsection{Abstention on Hard Examples}

In this scenario, we test with the labeler who abstains from labeling hard data, which are near to the true decision boundary.
This setting is common when the labeler wants to maximize the number of labels giving to the learner (e.g., in crowdsourcing where he is paid for each label provided).
The abstention pattern in this experiment is generated similarly to the previous scenario, except that the labeler abstains from labeling the examples having the smallest distances above instead of those with the largest distances.

Figure \ref{fig:set_easy_hard} (second row) shows the results for this scenario on the same 4 data sets above.
These results suggest that this is a more difficult setting for active learning.
From the figure, ALa and ALw are only better than the baselines when the abstention percentage is from 20-40\%.
For other abstention percentages, ALa, ALw, and ALg do not provide much advantage compared to PL.
However, when a good estimate of $r^*$ is available, ALa and ALw perform very well and are better than all the other algorithms.

{\bf Summary:}
The results above have shown that the proposed algorithms are useful for pool-based active learning with abstention feedbacks when the abstention percentage is within an appropriate range that depends on the problem.
The algorithms are especially useful when a good estimate of the abstention rate $r^*$ is available.
In practice, this estimate can be pre-computed from previous interactions between the learning systems and the labeler (e.g., using previous labeling preferences of the labeler), and then inputted into our algorithms as the priors $p_0[r]$. 
During the execution of our algorithms, this estimate will be gradually improved.

\section{Conclusion}
We proposed two new greedy algorithms under the Bayesian active learning with abstention feedbacks framework.
This framework is useful in many real-world scenarios, including learning from multiple labelers and under corrupted labels.
We proved that the algorithms have theoretical guarantees in the average and worst cases and also showed experimentally that they are useful for classification, especially when a good estimate of the abstention rate is available.
Our results suggest that keeping track and learning the abstention patterns of labelers are important for active learning with abstention feedbacks in practice.

\section*{Acknowledgments}

LSTH was supported by startup funds from Dalhousie University, the Canada Research Chairs program, the NSERC Discovery Grant RGPIN-2018-05447, and the NSERC Discovery Launch Supplement DGECR-2018-00181. BTN was supported by Vietnam National University Ho Chi Minh City (VNU-HCM) under grant number NCM2019-18-01 throughout this paper.

\bibliography{mybibfile}

\newpage
\appendix

\section{Proofs}

\subsection{Proof of Theorem \ref{theorem:near-opt-avg}}
\label{sec:proof-avg}
To prove this theorem, we first apply Theorem 5.2 in \cite{golovin2011adaptive}.
This requires us to prove that the utility function $g(S, (f,k))$ is adaptive monotone and adaptive submodular with respect to the joint prior distribution $q_0$.
Note that $g(S, (f,k))$ is the version space reduction function with respect to the joint prior $q_0$ on the joint space $\mathcal{F} \times \mathcal{K}$.
From the results in Section 9 of \cite{golovin2011adaptive}, version space reduction functions are adaptive monotone and adaptive submodular with respect to the corresponding prior.
Thus, the utility function $g(S, (f,k))$ is adaptive monotone and adaptive submodular with respect to the joint prior $q_0$.

With the above properties of $g$, applying Theorem 5.2 in \cite{golovin2011adaptive}, we have:
\[ G_{\text{avg}}(\pi_{\text{greedy}}) > (1 - 1/e) G_{\text{avg}}(\pi^*_{\text{avg}}), \]
where $\pi_{\text{greedy}}$ is the greedy algorithm that selects the examples maximizing the expected utility gain at each step.
From the proof of Theorem 4 of \cite{cuong2013active}, this greedy algorithm is equivalent to the maximum Gibbs error algorithm that selects the examples according to the criterion:
\begin{align}
\label{eq:gibbs}
x^* &= \arg \max_{x \in \mathcal{X}} \{ 1 - p_{i-1}[Z=1;x]^2 - \notag \\
&{\hskip 1.7cm} \sum_{y \in \mathcal{Y}} p_{i-1}[Y = y \wedge Z=0; x]^2 \},
\end{align}
where $p_{i-1}$ is the current posterior distribution, $Y$ is the random variable for the label of $x$, and $Z$ is the random variable for the abstention pattern of $x$.
To understand this equation, we can think of the considered problem as a classification problem with labels $(y,z=0)$ or $(z=1)$, where $(y,z=0)$ indicates an example is labeled the label $y$ and $(z=1)$ indicates an example is not labeled.

Since $y$ and $z$ are independent, Equation \eqref{eq:gibbs} is equivalent to:
\begin{eqnarray*}
x^* &=& \arg \max_{x \in \mathcal{X}} \{ 1 - p_{i-1}[Z=1;x]^2 - \\
&&{\hskip 0.5cm} \sum_{y \in \mathcal{Y}} (p_{i-1}[Z=0; x] ~ p_{i-1}[Y=y; x] )^2 \} \\
&=& \arg \max_{x \in \mathcal{X}} \{ 1 - p_{i-1}[Z=1;x]^2 - \\
&& {\hskip 0.5cm} p_{i-1}[Z=0; x]^2 \sum_{y \in \mathcal{Y}} p_{i-1}[Y=y; x]^2 \}.
\end{eqnarray*}

We also have:
\begin{eqnarray*}
p_{i-1}[Z=1;x] &=& \int p_{i-1}[r] ~ r(x) ~ dr \\
&=& \mathbb{E}_{r \sim p_{i-1}}[r(x)] \\
&=& \tilde{r}(x).
\end{eqnarray*}

Similarly, $p_{i-1}[Z=0;x] = 1 - \tilde{r}(x)$.

Hence, the previous equation is equivalent to:
\begin{equation*}
x^* = \arg \max_{x \in \mathcal{X}} \{ 1 - \tilde{r}(x)^2 - (1 - \tilde{r}(x))^2 \sum_{y \in \mathcal{Y}} p_{i-1}[Y=y; x]^2 \},
\end{equation*}
which is Equation \eqref{eq:criterion1}.
Therefore, the average-case BALAF algorithm is equivalent to $\pi_{\text{greedy}}$ and Theorem \ref{theorem:near-opt-avg} holds.

\subsection{Proof of Theorem \ref{theorem:near-opt-worst}}
\label{sec:proof-avg}

To prove this theorem, we first apply Theorem 3 in \cite{cuong2014near}.
This requires us to prove that the utility $g(S, (f,k))$ is pointwise monotone and pointwise submodular.
Note that $g(S, (f,k))$ is the version space reduction function with respect to the joint prior $q_0$ on the joint space $\mathcal{F} \times \mathcal{K}$.
From the proof of Theorem 5 in \cite{cuong2014near}, version space reduction functions are both pointwise monotone and pointwise submodular.
Thus, $g(S, (f,k))$ is pointwise monotone and pointwise submodular.

With the above properties of $g$, applying Theorem 3 in \cite{cuong2014near}, we have: 
\[ G_{\text{worst}}(\pi'_{\text{greedy}}) > (1 - 1/e) G_{\text{worst}}(\pi^*_{\text{worst}}), \]
where $\pi'_{\text{greedy}}$ is the greedy algorithm that selects the examples maximizing the worst-case utility gain at each step.
From the proof of Theorem 5 of \cite{cuong2014near}, this greedy algorithm is equivalent to the least confidence algorithm that selects the examples according to the criterion:
\begin{align}
\label{eq:lc}
x^* &= \arg \min_{x \in \mathcal{X}} \{ \max \{ p_{i-1}[Z=1;x], \notag \\ 
&{\hskip 2cm} \max_{y \in \mathcal{Y}} p_{i-1}[Y=y \wedge Z=0;x] \} \},
\end{align}
where $p_{i-1}$ is the current posterior distribution, $Y$ is the random variable for the label of $x$, and $Z$ is the random variable for the abstention pattern of $x$.
Similar to the proof of Theorem \ref{theorem:near-opt-avg} above, to understand this equation, we can think of the considered problem as a classification problem with labels $(y,z=0)$ or $(z=1)$, where $(y,z=0)$ indicates an example is labeled the label $y$ and $(z=1)$ indicates an example is not labeled.

Since $y$ and $z$ are independent, Equation \eqref{eq:lc} is equivalent to:
\begin{eqnarray*}
x^* &=& \arg \min_{x \in \mathcal{X}} \{ \max \{ p_{i-1}[Z=1;x], \\ 
&& {\hskip 0.5cm} \max_{y \in \mathcal{Y}} p_{i-1}[Y=y;x] ~ p_{i-1}[Z=0;x] \} \} \\
&=& \arg \min_{x \in \mathcal{X}} \{ \max \{ p_{i-1}[Z=1;x], \\ 
&& {\hskip 0.5cm} p_{i-1}[Z=0;x] \max_{y \in \mathcal{Y}} p_{i-1}[Y=y;x] \} \}.
\end{eqnarray*}

From the proof of Theorem \ref{theorem:near-opt-avg}, we have $p_{i-1}[Z=1;x] \allowbreak = \tilde{r}(x)$ and $p_{i-1}[Z=0;x] = 1 - \tilde{r}(x)$.

Hence, the previous equation is equivalent to:
\begin{equation*}
x^* = \arg \min_{x \in \mathcal{X}} \{ \max \{ \tilde{r}(x), (1 - \tilde{r}(x)) \max_{y \in \mathcal{Y}} p_{i-1}[Y=y;x] \} \},
\end{equation*}
which is Equation \eqref{eq:criterion2}.
Therefore, the worst-case BALAF algorithm is equivalent to $\pi'_{\text{greedy}}$ and Theorem \ref{theorem:near-opt-worst} holds.

\end{document}